\documentclass{article}
\usepackage[final]{corl_2024} 
\usepackage{graphicx}
\usepackage{subcaption}

\title{Cognitive Planning for Object Goal Navigation using Generative AI Models}

\author{
  Arjun P S\\
  IIT ISM  Dhanbad, India\\
  \And  
  Andrew Melnik\\
  Bielefeld University, Germany\\
  \And 
  Gora Chand Nandi\\
  IIIT Allahabad, India\\
}

\begin{document}
\maketitle

\begin{abstract}
     Recent advancements in Generative AI, particularly in Large Language Models (LLMs) and Large Vision-Language Models (LVLMs), offer new possibilities for integrating cognitive planning into robotic systems. In this work, we present a novel framework for solving the object goal navigation problem that generates efficient exploration strategies. Our approach enables a robot to navigate unfamiliar environments by leveraging LLMs and LVLMs to understand the semantic structure of the scene. To address the challenge of representing complex environments without overwhelming the system, we propose a 3D modular scene representation, enriched with semantic descriptions. This representation is dynamically pruned using an LLM-based mechanism, which filters irrelevant information and focuses on task-specific data. By combining these elements, our system generates high-level sub-goals that guide the robot’s exploration toward the target object. We validate our approach in simulated environments, demonstrating its ability to enhance object search efficiency while maintaining scalability in complex settings.
     Video Demonstration : \url{https://youtu.be/pvr1uaObL9M}
\end{abstract}


\begin{figure}[h]
        \centering
        \includegraphics[width=\textwidth]{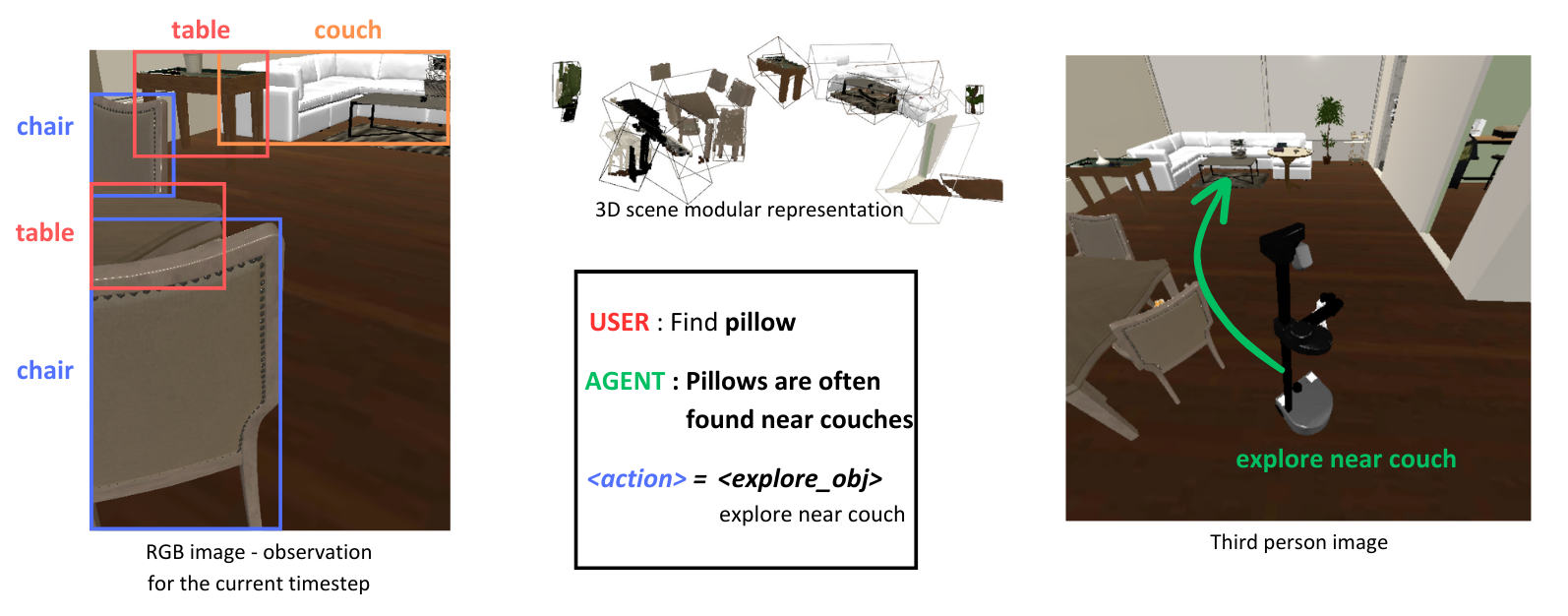}
        \caption{Overview of our framework with Homerobot Strech in Habitat simulation environment. The Robot in this episode is tasked to find a pillow. The agent, after considering all the objects in the scene (3D scene modular representation), decides to explore near the couch to find the pillow.}
        \label{fig:ovmm-pipelnie}
\end{figure}

\section{Introduction}

    Navigation in an unfamiliar environment to search for an object described in natural language is one of the most challenging problem in robotics \cite{yenamandra2024towards, melnik2023uniteam}. Even though these kind of task comes as second nature for us humans, the underlying process is really complex as it involves cognitive processing, using long term memory and experiences, and integrating the current sensory information with these processes. Achieving this with robots involve designing a language conditioned high-level planner \cite{mikami2024natural} that understands the semantic priors of this world and is able to devise intelligent sequential plans to efficiently explore the scene in search of the object. This is the core of an object goal navigation task. The agent, similar to a human being, must also have a cognitive high level-planner that can understand the semantic cues of our world, a low-level planner that can leverage on long term memory and experiences to act on the environment and an episodic memory to save critical information related to the current task. 

    Large Language Models (LLMs) and Large Vision Language Models (LVLMs) have shown the capability to comprehend the semantic priors of the world and reason about them from this understanding. This makes them an ideal candidate for a high level planner, that can take an action by reasoning about the scene. To plan efficiently in an environment by taking into account the underlying details of the scene, the agent should be provided with an efficient representation of the scene which is rich in semantic information but yet not overwhelmed with a mass of indistinguishable stimuli. Human beings, when exploring an unknown environment to find an object without a comprehensive map of the environment, follows a goal oriented approach in storing data \cite{melnik2018world}. This process involves a combination of Perception which process the incoming information, Attention which directs our focus to specific cues in the scene and memory which selectively encodes information that is perceived as important for achieving a specific goal. This aids in efficient use of cognitive resources and effective retrieval of this information when needed. This goal oriented approach enhances the chances of us finding the target object. We mimic this goal oriented approach with robots by using a combination of an Large language model for attending to relevant aspects, an large vision language model and an open-vocabulary image segmentation module for perceiving and processing the incoming information, and a 3D scene modular representation and a short term memory module for saving the processed information to memory.
    
    The agent when exploring the scene, procedurally builds a modular 3D scene representation, by using the processed information from the perception module. The 3D scene representation has nodes corresponding to objects of relevance in the scene, the relevancy of which is decided by a Large Language Model. The 3D scene representation generated is generally sparse, but becomes denser in the vicinity of the detected target object. This sparse to dense structure in encoding information mirrors the human cognition in strategically encoding information in a goal oriented task. The perception module involves an open vocabulary segmentation module which is responsible for identifying and segmenting all the objects in the scene and a Large Vision Language model used for describing the identified objects in natural language. This natural language description have details like the appearance of the object, the objects near it and the possible room this object is in. This information helps the high level planner to plan more effectively. The short term memory module temporarily holds and manipulate information needed for generating inferences about the target object.

\section{Related works}
    \subsection{LLM as a planner}
        Large language models (LLMs) with billions of parameters, trained on massive scale datasets have shown impressive generative capabilities with a generalized semantic understanding of the world. Many works like \citet{huang2022inner, ding2023task, song2023llmplanner, Lin2023, huang2022language, saycan2022arxiv} focused on grounding LLMs for high-level task planning with natural language task representation. \citet{huang2022language} worked on grounding high-level tasks expressed in natural language and decomposing them into low level plans and admissible actions. \citet{saycan2022arxiv} focused on grounding an LLM that provides high-level procedures for task completion, with value functions associated with these tasks.
    
    \subsection{Open Vocabulary Image Segmentation}
        Open vocabulary image segmentation refers to the recognition and delineation of an open category of objects in a frame. The flexibility of such a system to identify and segment objects forms a critical part in building an efficient representation of the scene. With the advent of powerful object detection and segmentation models like \citet{kirillov2023segment, wang2022yolov7, he2018mask, zhou2022detecting, zhao2023fast, ren2024grounded,liu2023grounding}, the agent can detect and segment the objects in the scene. Many open vocabulary segmentation modules like \citet{zhou2022detecting, ren2024grounded, kirillov2023segment} require the labels of the objects to perform instance segmentation. Image tagging models like \citet{zhang2023recognize} can recognize the objects in a scene and return their labels, which can be used as text prompt to these models.
    
    \subsection{Representating the environment for task planning}
        Prior works like \citet{conceptgraphs, chen2022nlmapsaycan, huang23vlmaps, chaplot2020object, zhang20233daware, chen2023object} focused on building efficient representations of the environment to facilitate downstream task planning. The representations generated can be used for understanding the semantics of the scene, thus allowing the agent to query the environment for an object using natural language \citet{conceptgraphs, chen2022nlmapsaycan} and generate task level plans \citet{rana2023sayplan, chen2022nlmapsaycan} or low-level actions \citet{chen2023object}. Even though many scene representations like \citet{chaplot2020object, chen2023object} have created a scalable and efficient representation that can be used for navigation, it lacks contextual information about the scene, which is invaluable when exploring a scene. Scene representations like \citet{conceptgraphs, chen2022nlmapsaycan} have a really dense representation, which is often not required when exploring an unfamiliar environment in search of a target object $G_o$. Our present research is inspired by the human way of cognition with an affordance based memory that attends and stores only those information which are relevant to the task at hand. We leverage on the ability of In-context learning in LLMs to prune out irrelevant information and then using an LVLM to reason about the relevant pruned list of objects, thus creating a sparse as well as information rich representation of the environment. The scene representation near the target object $G_o$ is made denser so as to facilitate further downstream task planning like manipulation of the object.
    
    \subsection{Object goal navigation}
        Object goal navigation task involves finding a target object $G_o$ in an unfamiliar environment. Efficient exploration through the environment requires deep understanding of semantic priors of the world. Prior works like \citet{chaplot2020object, alhalah2022zero, mousavian2019visual, chang2020semantic, liang2021sscnav, wu2018learning} tried to learn these semantic cues from egocentric RGB and depth images \citet{alhalah2022zero, mousavian2019visual}, semantic map \citet{chaplot2020object} and even from YouTube videos \citet{chang2020semantic}. These learned semantic cues might not generalize well to new unseen environments. \citet{alhalah2022zero} focused on using a modular transfer learning model to generalize the learned policy for a particular task to multiple tasks and \citet{wortsman2019learning} used meta learning to generalize to unfamiliar environments. LLMs on the other hand exhibit extraordinary contextual awareness and ability to understand the semantic cues of our world. We leverage on this capability to efficiently explore unseen environments for the target object $G_o$.
    
        Recent works like \citet{rajvanshi2023saynav, dorbala2023embodied, shah2023navigation} have leveraged the capability of LLMs to understand the semantic priors for object goal navigation tasks. \citet{rajvanshi2023saynav} leveraged the planning capabilities of LLMs to devise a sequential plan that includes point goals and target state information for executing multiple object goal navigation tasks in parallel. These point goals generated by an LLM is fulfilled through a low level(execution level) controller. \citet{shah2023navigation} focused on using the semantic predictions from LLMs as a heuristic for a frontier based exploration strategy to find the target object $G_o$. \citet{dorbala2023embodied} focused on solving language-driven zero-shot object goal navigation problem \citet{majumdar2023zson} by using an LLM to navigate to the target object $G_o$, given a natural language description of it. 
        
        Most of these approaches either focuses on building a semantically rich map by exploring the environment first, and then utilizes a policy to navigate on top of this representation or have a procedurally built semantic map which lacks rich semantic information but is built on the fly and a policy trained on top of this, which learns the semantic priors of the world. Such policies, especially those trained on simulators lack generalization capabilities. We utilize the generalization capabilities of foundational models to understand and reason about the context through natural language, to procedurally build a 3D modular representation which encodes rich semantic data, and generate plans for an object goal navigation task.

\section{METHOD}
\begin{figure*}
    \centering
    \includegraphics[width=\textwidth]{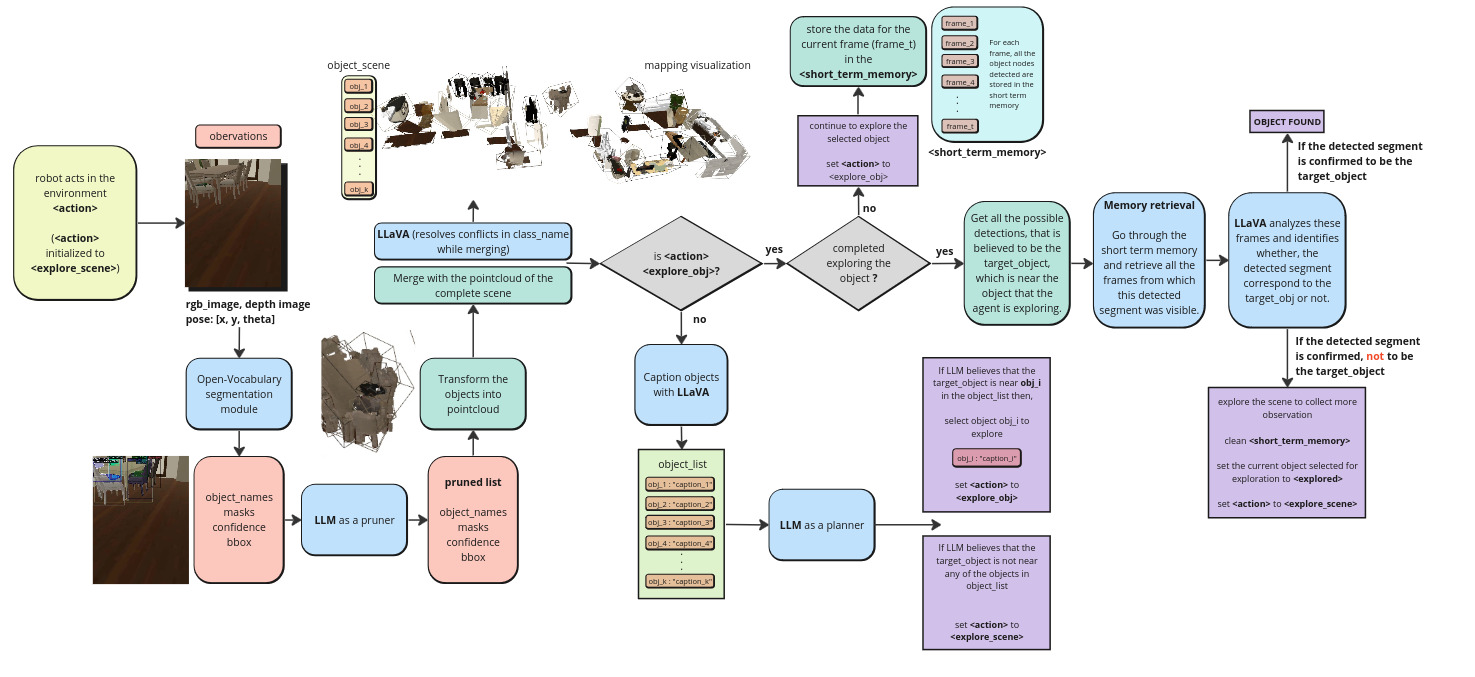}
    \caption{Architecture of the proposed pipeline.The agent explores the environment and collects observations (RGBD image and Pose). An open-vocabulary segmentation module is used to identify the objects in the current frame. Pruner takes in these detected segments and prunes out unwanted segments. The pruned segments are either initialized as a new node in the 3D scene representation or merged with an existing one, based on a similarity criteria. All new nodes are captioned with LLaVA, to provide semantic information to the LLM based planner. The agent then chooses a node to explore closer, to find the target object $G_o$. While doing so, the agent stores frame wise information in the short term memory module. If the agent decides that no objects in the 3D scene representation has a good chance of finding the target object $G_o$ closer to it, it continues to explore the scene and build the 3D scene representation.}
    \label{fig:ovmm-architecture}
\end{figure*}

An overview of our approach is given in figure \ref{fig:ovmm-architecture}. The agent receives RGB image, depth image, base and camera pose as inputs in every timestep. An open vocabulary semantic segmentation module identifies objects in the RGB image and generates mask for the same. LLM acting as a pruner identifies the most important objects in the scene in regards to their utility in understanding the semantic priors of the environment and prunes the remaining detections. These detected segments are then converted into a 3D scene representation, where each object is signified by a node. An object node stores information regarding the position of the object, pointcloud, frame in which this object was detected, mask and label of this object in that frame, and a semantic description of the object in the scene. LLM, grounded for an object goal navigation task, uses the 3D scene modular representation to decide whether to continue exploring the environment or to move closer to an already detected object in the scene that has a high probability of finding the target object $G_o$. The agent triggers its short term memory module and stores information for every frame if it decides to explore closer to an object. Upon reaching the object, the agent utilize the detections from the short term memory, to construct hypotheses regarding the target object.

\subsection{Open vocabulary image segmentation}
The open vocabulary segmentation module segments the RGB frame $I_t$ at the current timestep $t$, given the natural language description of the object. Given an RGB image $I_t$ of the scene, we use a combination of Recognize anything model(RAM) \citet{zhang2023recognize}, Grounding Dino and FastSAM \citet{ren2024grounded} for open vocabulary semantic segmentation. RAM is a strong foundational model used for image tagging, it is tasked with finding the object tags $c_t^i$ in the scene. The set of object tags $C_t = (c_t^1, c_t^2, ..., c_t^m)$ is then passed as a text prompt $T_t$ $(= C_t)$ to Grounding-Dino \citet{liu2023grounding}, which detects the presence of classes $C_t$ in the scene. Grounding-Dino returns the class labels $(l_t^1, l_t^2, ..., l_t^m)$ of objects in the scene, confidence of the predictions $(p_t^1, p_t^2, ..., p_t^m)$ and their bounding boxes $(b_t^1, b_t^2, ..., b_t^m)$. The detected classes in the scene $(l_t^1, l_t^2, ..., l_t^m)$ are then passed as box prompt $(b_t^1, b_t^2, ..., b_t^m)$ through FastSAM \citet{zhao2023fast} to get their corresponding masks $(m_t^1, m_t^2, ..., m_t^m)$.

\subsection{LLM as a pruner}
We leverage the In-context learning abilities of an LLM to prune the detected class labels $(l_t^1, l_t^2, ..., l_t^m)$ generated by the open vocabulary segmentation module for the timestep $t$. In-context learning is a paradigm that empowers pre-trained LLMs to ground to a particular task without fine-tuning the model, by providing the LLM with task level demonstrations along with the prompt. The language model here is provided with task-specific examples containing a pair of input $P_i$ and output $P_o$ object sets. The input set $P_i$ contains class labels $(l^1, l^2, ..., l^m)$ and the output set $P_o (\subseteq P_i)$, contains those objects that have a higher utility when it comes to understanding the semantic priors of the scene. The text prompt provided to the LLM to achieve this behaviour is given in fig \ref{fig:pruner}.

\subsection{3D scene modular representation}
The pruned semantic masks $(m_t^1, m_t^2, ..., m_t^m)$ and labels $(l_t^1, l_t^2, ..., l_t^m)$ from the open vocabulary segmentation module, along with the depth image $I_t^D$ and camera pose $P_t$ in the current time step $t$ are used to generate 3D object nodes $(N_{c1}, N_{c2}, ..., N_{ck})$ for each object $O_i$ in that frame $(I_t, I_t^D)$ using a modified concept graphs pipeline \citet{conceptgraphs}. In concept graphs, the 3D object nodes $N_{ci}$ in the scene representation are fused from multiple views considering their spatial overlap, using an object association strategy that compares the semantic and visual similarity of these objects. Merging objects from different frames often creates conflicts in class labels, as the object segment can be associated with different class labels in different frames. We propose to use a Large Vision Language Model (LVLM) to resolve this conflict. The LVLM is tasked to find the class label that best suits the object detected, by giving it a cropped image of the detected object and the conflicting class labels. 

Considering an object goal navigation task, where the agent is tasked to find a target object $G_o$ in the scene, further pruning of the concept graphs representation considering the utility of objects, in the regions not in the immediate relevant environment of the target object $G_o$, so as to generate a really sparse representation will improve scalability, which enables the system to handle larger and more complex environments without a proportional increases in computational or memory requirements. But having a sparse representation around the target object, constrains the agent in generating downstream plans for manipulating this object. To resolve this, the agent is conditioned to generate a really dense and semantically rich representation around the target object $G_o$ which will enable the agent to execute downstream planning and manipulation tasks. 

The object nodes generated are captioned using an LVLM in order to extract more information regarding the semantics of the object. Node captioning system in concept graphs is done while generating the scene graph, after construction of the entire 3D scene representation. This restricts the robots in fully utilizing the information while generating plans to explore the environment. But generating captions on the fly utilizes a lot of compute power, thus slowing down the robot by a considerable factor. To resolve this, the agent is tasked to caption nodes only when the action at the current timestep is $<explore\textunderscore scene>$, i.e. the agent captions the node in every timestep, until it selects an object to explore further. In the phase where the agent explores towards a selected object ($action = <explore\textunderscore obj>$), it saves all the relevant information regarding objects. The agent processes this information and generates captions for the object nodes at the end of this phase which is indicated when the robot reaches the selected object node.
\begin{figure*}
    \centering
    \begin{subfigure}{0.45\textwidth}
        \centering
        \includegraphics[width=\linewidth]{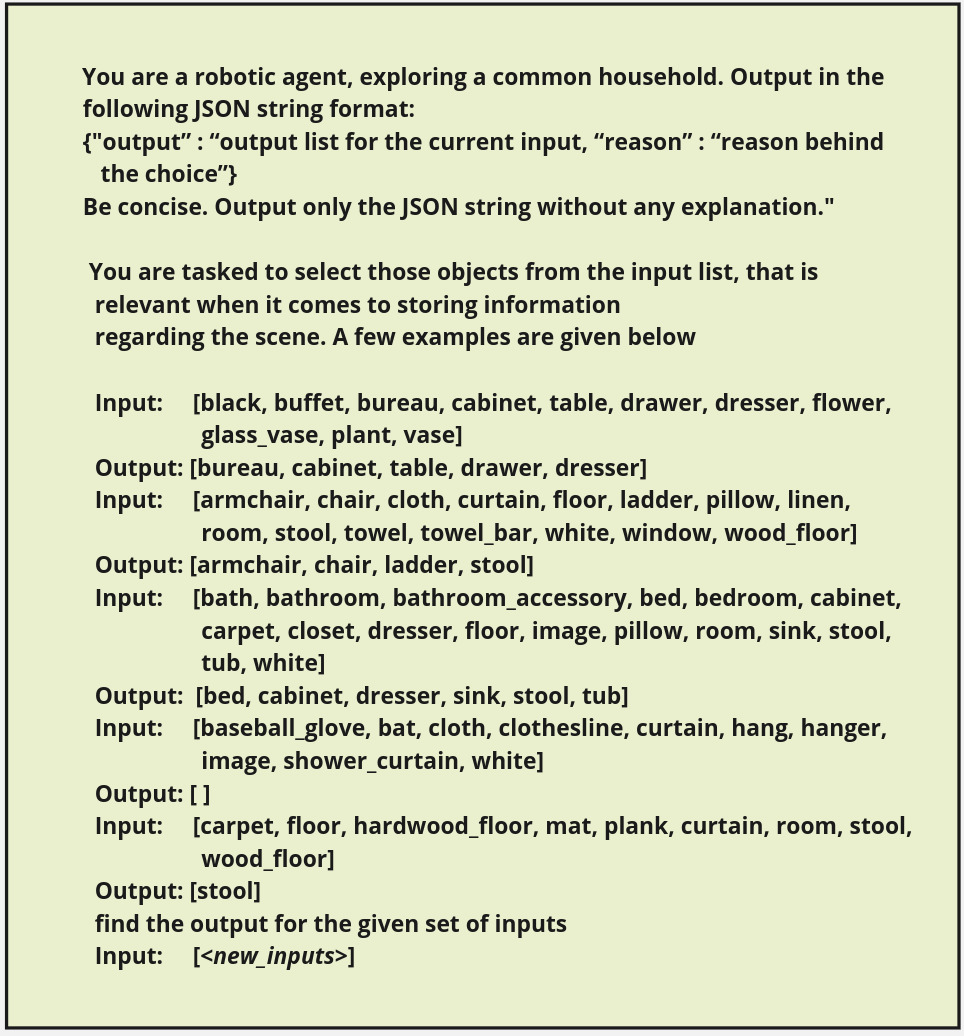}
        \caption{Prompt to LLM for pruning the detections}
    \end{subfigure}\hfill
    \begin{subfigure}{0.45\textwidth}
        \centering
        \includegraphics[width=\linewidth]{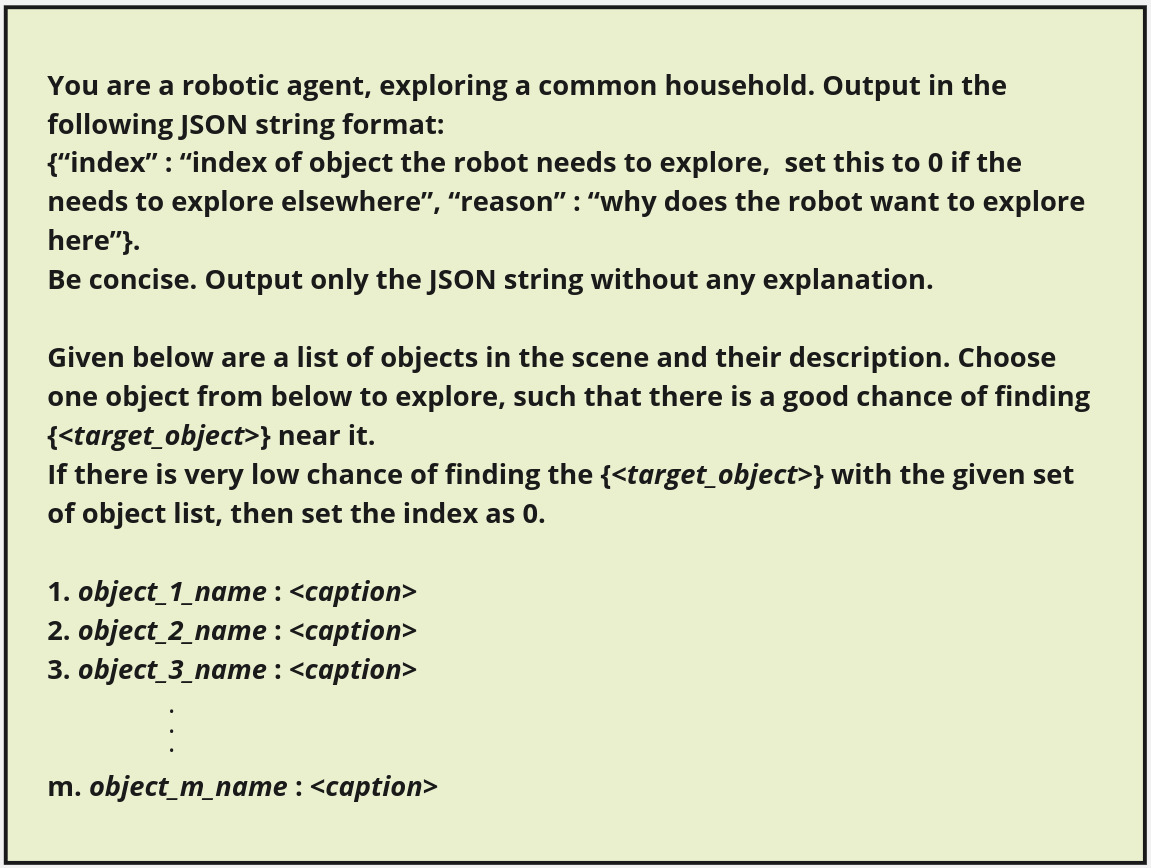}
        \caption{Prompt to LLM for planning.}
    \end{subfigure}
\end{figure*}
\subsection{LLM as a planner}
LLM, grounded in the current task using task level description of the scene is used to generate a high level plan for the agent. Given the current objects in the environment (objects in the constructed 3D scene representation ($N_{c1}, N_{c2}, ..., N_{ck}$)) and the generated descriptions for these object nodes, the LLM is tasked to identify whether to continue exploring the environment randomly and build up the 3D scene representation, or to explore the regions closer to an object node $N_{cj}$. If the LLM decides to explore the scene more, then $action$ is set to $<explore\textunderscore scene>$ and if it believes that there is a high chance of finding the target object $G_o$ near any of the detected objects, then $action$ is set to $<explore\textunderscore obj>$. This ability of an LLM to extract semantic relationships between the target object $G_o$ and objects in the scene ($N_{c1}, N_{c2}, ..., N_{ck}$) helps the agent to explore efficiently in the environment. The text prompt provided to the LLM to achieve this behaviour is given in fig \ref{fig:planner}.

\subsection{Short term memory(STM) and reasoning}

Short term memory(STM) is a temporary storage module for holding and manipulating information amidst task execution. This working memory module is triggered when the $action$ is set to $<explore\textunderscore obj>$. The agent stores processed information (inference by the open vocabulary segmentation module and segment in that frame which contributed to object nodes in the 3D scene representation) for all the frames and holds it, until it has completely explored near an object. Having a short term memory (temporary) oven an long term or permanent memory reduces memory constraints and increases the scalability of the architecture. 

Upon reaching the target object, the agent pans around its axis to collect information regarding the scene, which is processed by the segmentation module primed with the target object. This generates all the possible segments that can be regarded as the target object. The agent then retrieves all the frames in the short term memory, from which these segments are visible, thus generating different views for the same detected segment. This is done by finding the spatial overlap of pointclouds between the segment of interest and the objects from frames stored in the short term memory.

The agent then tasks an LVLM (LLaVA \citet{liu2023llava, liu2023improvedllava}) to verify whether the detected segment in all the views is the target object or not. The determination of whether the detected segment represents the target object $G_o$ relies on the percentage of frames (different views) in which the LVLM (LLaVA) confirms the presence of the target object, serving as the decisive criterion for the final decision. If none of the detected segments are regarded as the target object then the current object selected for exploring is set to explored, the short term memory is cleared and the agent's $action$ is set to $<explore\textunderscore scene>$.
\subsection{Execution level planner}

We make use of a goal oriented and episodic semantic mapping module \citet{chaplot2020object}, to build an episodic 2D scene obstacle map, frontier map and a goal map. In the phase where $action$ is $<explore\textunderscore scene>$, the goal map is same as that of the frontier map and when the $action$ is $<explore\textunderscore obj>$, the goal map is constructed in such a way that the robot can navigate to a region close to the object selected by the LLM. The goal map and the obstacle map is then parsed to a fast marching method planner, which plans execution level actionable sub-goals to reach the goal.

\section{EXPERIMENT AND RESULTS}
We evaluate our stack to explore for a target object by building a 3D scene representation of the environment on the HomeRobot: Open Vocabulary Mobile Manipulation \citet{homerobotovmm} simulation benchmark. This simulation benchmark have multi-room realistic environments, with a diverse and complex set of receptacles (eg: chair, table, couch, bed, toilet etc.) and articles (eg: pen, book, cellphone, tray, apple etc.) that can be manipulated.The environments, designed in Habitat \citet{puig2023habitat3, szot2021habitat, habitat19iccv} provide a cluttered and navigable scene having multiple rooms with manipulable objects placed on top of receptacles, to test various benchmarks related to navigation and manipulation of objects in unfamiliar environments. We chose this dataset over conventional object-nav datasets and benchmarks, because of the availability of a wide range of small objects, that are manipulable. The main focus of our work is to explore for these manipulable objects. However, the dataset provided by this benchmark does not consider the general semantics of the world while spawning objects on top of receptacles. For example, there is an episode in this dataset that spawns in an apple on top of a toilet. We went through the dataset and curated custom episodes that follow these semantics to test the stack.

We test our stack for exploration by spawning the robot in a random location in the scene and asking it to find the target object within 500 steps. We evaluate the agents performance in navigating to the target object by considering success rate (SR) and success rate weighted by path length (SPL) which indicates the efficiency of the planner. The focus of our approach is in creating a planning pipeline that mimics human cognition in the way we perceive, store, express and regulate information. To test the efficacy of the approach compared to human level cognition, we tasked a set of random human volunteers to find the target object, for the same set of episodes that were given to the LLM based planner and  measured the success rate (SR) and the success rate weighted by path length (SPL). The results of the experiments involving human and LLM agent is given in table \ref{table_res}.

\begin{table}[ht]
    \centering
    \begin{minipage}{0.45\linewidth}
        \centering
        \caption{Experimental results}
        \label{table_res}
        \begin{tabular}{|c||c||c|}
            \hline
            \textbf{AGENT} & \textbf{SR} & \textbf{SPL}\\
            \hline
            \textbf{Human} & 0.9375 & 0.759\\
            \hline
            \textbf{GPT-4} & 0.4375 & 0.272\\
            \hline
            \textbf{GPT-3.5 Turbo} & 0.0 & 0.0\\
            \hline
        \end{tabular}
    \end{minipage}%
    \hspace{0.5cm}
    \begin{minipage}{0.45\linewidth}
        \centering
        \caption{Ablation analysis}
        \label{table_abl}
        \begin{tabular}{|c||c||c|}
            \hline
            \textbf{AGENT} & \textbf{SR} & \textbf{SPL}\\
            \hline
            \textbf{GPT-4} & 0.437 & 0.272\\
            \hline
            \textbf{GPT-4 w/o STM} & 0.125 & 0.089\\
            \hline
            \textbf{GPT-4 w/o pruner} & 0.182 & 0.113\\
            \hline
            \textbf{GPT-4 w/o captions} & 0.364 & 0.225\\
            \hline
        \end{tabular}
    \end{minipage}
\end{table}

\section{DISCUSSION}
\subsection{Comparison with human cognition}
Experimental results from table \ref{table_res}, shows the comparison between human agent and LLM agent in completing an object goal navigation task. Although GPT-4 based agent's performance is far from human level performance, it has exhibited a similar thought process in exploring near certain objects in the scene, for finding the target object $G_o$. GPT-3.5 on the other hand failed to reason in most of the episodes, leading to an inefficient and meaningless exploration of the scene. Most of the failure scenarios with GPT-3.5 was when it failed to find the target object within the 500 step limit.

\subsection{Ablation studies}
To signify the importance of having a short term memory based inference module, object captioning module and a detection pruner, we performed ablation analysis with these, the results of which are presented in table \ref{table_abl}. 

GPT-4 with the short term memory(STM) based inference module, more effectively identified the target object $G_o$, when compared to the agent without this module. The strategy of finding the same detected segment from multiple views, already encountered while exploring and constructing a hypothesis from these views, improved the agents performance in identifying the target object. Comparing with an agent with STM, Most of the failure scenario for the agent without STM was due to false positive detection of target object $G_o$, leading to premature termination.

LLM based pruner is tasked to prune out smaller objects in an environment like pen, book, pillow etc, and keep object like chair, kitchen table, bed etc, which are important in understanding the semantics of the scene. By not doing so, the agent is flooded with a lot of redundant information, which can lead to inefficient plans. Table \ref{table_abl} shows the performance of the agent without the pruner. The agent without a pruning module engages in a lot of inefficient exploration, which can be seen with the reduced SPL, when compared to an agent with pruner.

Captioning nodes can capture a lot of semantic information, which can increase the efficiency of the planner. GPT-4 based agent, provided with captions performed better at reasoning about the target object $G_o$ being near certain objects in the scene, leading to a more optimal path when compared to the case when the agent was only provided with the object names. Object captions can capture the semantic differences between objects of the same label. For example, the chance of finding an apple is more near a table in the kitchen than in the bedroom. Compared to the agent that had access to object captions, most of the failure scenarios transpired when the agent was not able to find the object within the 500 step limit. This can be attributed to the inefficient exploration due a lack of semantic understanding of the scene.

\subsection{Limitations}
Regardless of its capabilities, this framework possesses certain limitations, with object detection and segmentation being the most critical among them, creating false positive detection of objects. Another limitation was with the object captioning system, which often lead to exploration of unwanted objects. This can be attributed to the current limitations of LVLMs like LLaVA. Using a proprietary LLM like GPT-4 has a lot of challenges associated with it like the economic cost and slower control loop as its a cloud hosted model.

\section{CONCLUSION}
In this paper we introduced a framework to tackle the object goal navigation problem, by leveraging the generalization and semantic reasoning capabilities of Large Language Models and Large Vision Language Models to generate intelligent plans, by utilizing an efficient 3D scene representation of the environment. Future works may also explore the use of a more efficient representation, which the agent can use to reason about the environment. 

\clearpage
\acknowledgments{This work was supported by DAAD PPP-India program under project number 57622912. This work was supported by Ministry of Culture and Science of the State of North Rhine-Westphalia, Germany, through the KI-Starter funding program.}

\bibliography{mainbib}

\begin{thebibliography}{41}
\providecommand{\natexlab}[1]{#1}
\providecommand{\url}[1]{\texttt{#1}}
\expandafter\ifx\csname urlstyle\endcsname\relax
  \providecommand{\doi}[1]{doi: #1}\else
  \providecommand{\doi}{doi: \begingroup \urlstyle{rm}\Url}\fi

\bibitem[Yenamandra et~al.(2024)Yenamandra, Ramachandran, Khanna, Yadav, Vakil, Melnik, B{\"u}ttner, Harz, Brown, Nandi, et~al.]{yenamandra2024towards}
S.~Yenamandra, A.~Ramachandran, M.~Khanna, K.~Yadav, J.~Vakil, A.~Melnik, M.~B{\"u}ttner, L.~Harz, L.~Brown, G.~C. Nandi, et~al.
\newblock Towards open-world mobile manipulation in homes: Lessons from the neurips 2023 homerobot open vocabulary mobile manipulation challenge.
\newblock \emph{arXiv preprint arXiv:2407.06939}, 2024.

\bibitem[Melnik et~al.(2023)Melnik, B{\"u}ttner, Harz, Brown, Nandi, PS, Yadav, Kala, and Haschke]{melnik2023uniteam}
A.~Melnik, M.~B{\"u}ttner, L.~Harz, L.~Brown, G.~C. Nandi, A.~PS, G.~K. Yadav, R.~Kala, and R.~Haschke.
\newblock Uniteam: Open vocabulary mobile manipulation challenge.
\newblock \emph{arXiv preprint arXiv:2312.08611}, 2023.

\bibitem[Mikami et~al.(2024)Mikami, Melnik, Miura, and Hautam{\"a}ki]{mikami2024natural}
Y.~Mikami, A.~Melnik, J.~Miura, and V.~Hautam{\"a}ki.
\newblock Natural language as polices: Reasoning for coordinate-level embodied control with llms.
\newblock \emph{arXiv preprint arXiv:2403.13801}, 2024.

\bibitem[Melnik et~al.(2018)Melnik, Sch{\"u}ler, Rothkopf, and K{\"o}nig]{melnik2018world}
A.~Melnik, F.~Sch{\"u}ler, C.~A. Rothkopf, and P.~K{\"o}nig.
\newblock The world as an external memory: The price of saccades in a sensorimotor task.
\newblock \emph{Frontiers in behavioral neuroscience}, 12:\penalty0 253, 2018.

\bibitem[Huang et~al.(2022)Huang, Xia, Xiao, Chan, Liang, Florence, Zeng, Tompson, Mordatch, Chebotar, Sermanet, Brown, Jackson, Luu, Levine, Hausman, and Ichter]{huang2022inner}
W.~Huang, F.~Xia, T.~Xiao, H.~Chan, J.~Liang, P.~Florence, A.~Zeng, J.~Tompson, I.~Mordatch, Y.~Chebotar, P.~Sermanet, N.~Brown, T.~Jackson, L.~Luu, S.~Levine, K.~Hausman, and B.~Ichter.
\newblock Inner monologue: Embodied reasoning through planning with language models.
\newblock In \emph{arXiv preprint arXiv:2207.05608}, 2022.

\bibitem[Ding et~al.(2023)Ding, Zhang, Paxton, and Zhang]{ding2023task}
Y.~Ding, X.~Zhang, C.~Paxton, and S.~Zhang.
\newblock Task and motion planning with large language models for object rearrangement, 2023.

\bibitem[Song et~al.(2023)Song, Wu, Washington, Sadler, Chao, and Su]{song2023llmplanner}
C.~H. Song, J.~Wu, C.~Washington, B.~M. Sadler, W.-L. Chao, and Y.~Su.
\newblock Llm-planner: Few-shot grounded planning for embodied agents with large language models, 2023.

\bibitem[Lin et~al.(2023)Lin, Agia, Migimatsu, Pavone, and Bohg]{Lin2023}
K.~Lin, C.~Agia, T.~Migimatsu, M.~Pavone, and J.~Bohg.
\newblock Text2motion: from natural language instructions to feasible plans.
\newblock \emph{Autonomous Robots}, Nov 2023.
\newblock ISSN 1573-7527.
\newblock \doi{10.1007/s10514-023-10131-7}.

\bibitem[Huang et~al.(2022)Huang, Abbeel, Pathak, and Mordatch]{huang2022language}
W.~Huang, P.~Abbeel, D.~Pathak, and I.~Mordatch.
\newblock Language models as zero-shot planners: Extracting actionable knowledge for embodied agents.
\newblock \emph{arXiv preprint arXiv:2201.07207}, 2022.

\bibitem[Ahn et~al.(2022)Ahn, Brohan, Brown, Chebotar, Cortes, David, Finn, Fu, Gopalakrishnan, Hausman, Herzog, Ho, Hsu, Ibarz, Ichter, Irpan, Jang, Ruano, Jeffrey, Jesmonth, Joshi, Julian, Kalashnikov, Kuang, Lee, Levine, Lu, Luu, Parada, Pastor, Quiambao, Rao, Rettinghouse, Reyes, Sermanet, Sievers, Tan, Toshev, Vanhoucke, Xia, Xiao, Xu, Xu, Yan, and Zeng]{saycan2022arxiv}
M.~Ahn, A.~Brohan, N.~Brown, Y.~Chebotar, O.~Cortes, B.~David, C.~Finn, C.~Fu, K.~Gopalakrishnan, K.~Hausman, A.~Herzog, D.~Ho, J.~Hsu, J.~Ibarz, B.~Ichter, A.~Irpan, E.~Jang, R.~J. Ruano, K.~Jeffrey, S.~Jesmonth, N.~Joshi, R.~Julian, D.~Kalashnikov, Y.~Kuang, K.-H. Lee, S.~Levine, Y.~Lu, L.~Luu, C.~Parada, P.~Pastor, J.~Quiambao, K.~Rao, J.~Rettinghouse, D.~Reyes, P.~Sermanet, N.~Sievers, C.~Tan, A.~Toshev, V.~Vanhoucke, F.~Xia, T.~Xiao, P.~Xu, S.~Xu, M.~Yan, and A.~Zeng.
\newblock Do as i can and not as i say: Grounding language in robotic affordances.
\newblock In \emph{arXiv preprint arXiv:2204.01691}, 2022.

\bibitem[Kirillov et~al.(2023)Kirillov, Mintun, Ravi, Mao, Rolland, Gustafson, Xiao, Whitehead, Berg, Lo, Dollár, and Girshick]{kirillov2023segment}
A.~Kirillov, E.~Mintun, N.~Ravi, H.~Mao, C.~Rolland, L.~Gustafson, T.~Xiao, S.~Whitehead, A.~C. Berg, W.-Y. Lo, P.~Dollár, and R.~Girshick.
\newblock Segment anything, 2023.

\bibitem[Wang et~al.(2022)Wang, Bochkovskiy, and Liao]{wang2022yolov7}
C.-Y. Wang, A.~Bochkovskiy, and H.-Y.~M. Liao.
\newblock Yolov7: Trainable bag-of-freebies sets new state-of-the-art for real-time object detectors, 2022.

\bibitem[He et~al.(2018)He, Gkioxari, Dollár, and Girshick]{he2018mask}
K.~He, G.~Gkioxari, P.~Dollár, and R.~Girshick.
\newblock Mask r-cnn, 2018.

\bibitem[Zhou et~al.(2022)Zhou, Girdhar, Joulin, Kr{\"a}henb{\"u}hl, and Misra]{zhou2022detecting}
X.~Zhou, R.~Girdhar, A.~Joulin, P.~Kr{\"a}henb{\"u}hl, and I.~Misra.
\newblock Detecting twenty-thousand classes using image-level supervision.
\newblock In \emph{ECCV}, 2022.

\bibitem[Zhao et~al.(2023)Zhao, Ding, An, Du, Yu, Li, Tang, and Wang]{zhao2023fast}
X.~Zhao, W.~Ding, Y.~An, Y.~Du, T.~Yu, M.~Li, M.~Tang, and J.~Wang.
\newblock Fast segment anything, 2023.

\bibitem[Ren et~al.(2024)Ren, Liu, Zeng, Lin, Li, Cao, Chen, Huang, Chen, Yan, Zeng, Zhang, Li, Yang, Li, Jiang, and Zhang]{ren2024grounded}
T.~Ren, S.~Liu, A.~Zeng, J.~Lin, K.~Li, H.~Cao, J.~Chen, X.~Huang, Y.~Chen, F.~Yan, Z.~Zeng, H.~Zhang, F.~Li, J.~Yang, H.~Li, Q.~Jiang, and L.~Zhang.
\newblock Grounded sam: Assembling open-world models for diverse visual tasks, 2024.

\bibitem[Liu et~al.(2023)Liu, Zeng, Ren, Li, Zhang, Yang, Li, Yang, Su, Zhu, and Zhang]{liu2023grounding}
S.~Liu, Z.~Zeng, T.~Ren, F.~Li, H.~Zhang, J.~Yang, C.~Li, J.~Yang, H.~Su, J.~Zhu, and L.~Zhang.
\newblock Grounding dino: Marrying dino with grounded pre-training for open-set object detection, 2023.

\bibitem[Zhang et~al.(2023)Zhang, Huang, Ma, Li, Luo, Xie, Qin, Luo, Li, Liu, et~al.]{zhang2023recognize}
Y.~Zhang, X.~Huang, J.~Ma, Z.~Li, Z.~Luo, Y.~Xie, Y.~Qin, T.~Luo, Y.~Li, S.~Liu, et~al.
\newblock Recognize anything: A strong image tagging model.
\newblock \emph{arXiv preprint arXiv:2306.03514}, 2023.

\bibitem[Gu et~al.(2023)Gu, Kuwajerwala, Morin, Jatavallabhula, Sen, Agarwal, Rivera, Paul, Ellis, Chellappa, Gan, {de Melo}, Tenenbaum, Torralba, Shkurti, and Paull]{conceptgraphs}
Q.~Gu, A.~Kuwajerwala, S.~Morin, K.~Jatavallabhula, B.~Sen, A.~Agarwal, C.~Rivera, W.~Paul, K.~Ellis, R.~Chellappa, C.~Gan, C.~{de Melo}, J.~Tenenbaum, A.~Torralba, F.~Shkurti, and L.~Paull.
\newblock Conceptgraphs: Open-vocabulary 3d scene graphs for perception and planning.
\newblock \emph{arXiv}, 2023.

\bibitem[Chen et~al.(2022)Chen, Xia, Ichter, Rao, Gopalakrishnan, Ryoo, Stone, and Kappler]{chen2022nlmapsaycan}
B.~Chen, F.~Xia, B.~Ichter, K.~Rao, K.~Gopalakrishnan, M.~S. Ryoo, A.~Stone, and D.~Kappler.
\newblock Open-vocabulary queryable scene representations for real world planning.
\newblock In \emph{arXiv preprint arXiv:2209.09874}, 2022.

\bibitem[Huang et~al.(2023)Huang, Mees, Zeng, and Burgard]{huang23vlmaps}
C.~Huang, O.~Mees, A.~Zeng, and W.~Burgard.
\newblock Visual language maps for robot navigation.
\newblock In \emph{Proceedings of the IEEE International Conference on Robotics and Automation (ICRA)}, London, UK, 2023.

\bibitem[Chaplot et~al.(2020)Chaplot, Gandhi, Gupta, and Salakhutdinov]{chaplot2020object}
D.~S. Chaplot, D.~Gandhi, A.~Gupta, and R.~Salakhutdinov.
\newblock Object goal navigation using goal-oriented semantic exploration.
\newblock In \emph{In Neural Information Processing Systems}, 2020.

\bibitem[Zhang et~al.(2023)Zhang, Dai, Meng, Fan, Chen, Xu, and Wang]{zhang20233daware}
J.~Zhang, L.~Dai, F.~Meng, Q.~Fan, X.~Chen, K.~Xu, and H.~Wang.
\newblock 3d-aware object goal navigation via simultaneous exploration and identification, 2023.

\bibitem[Chen et~al.(2023)Chen, Chabal, Laptev, and Schmid]{chen2023object}
S.~Chen, T.~Chabal, I.~Laptev, and C.~Schmid.
\newblock Object goal navigation with recursive implicit maps, 2023.

\bibitem[Rana et~al.(2023)Rana, Haviland, Garg, Abou-Chakra, Reid, and Suenderhauf]{rana2023sayplan}
K.~Rana, J.~Haviland, S.~Garg, J.~Abou-Chakra, I.~Reid, and N.~Suenderhauf.
\newblock Sayplan: Grounding large language models using 3d scene graphs for scalable task planning.
\newblock In \emph{7th Annual Conference on Robot Learning}, 2023.

\bibitem[Al-Halah et~al.(2022)Al-Halah, Ramakrishnan, and Grauman]{alhalah2022zero}
Z.~Al-Halah, S.~K. Ramakrishnan, and K.~Grauman.
\newblock Zero experience required: Plug and play modular transfer learning for semantic visual navigation, 2022.

\bibitem[Mousavian et~al.(2019)Mousavian, Toshev, Fiser, Kosecka, Wahid, and Davidson]{mousavian2019visual}
A.~Mousavian, A.~Toshev, M.~Fiser, J.~Kosecka, A.~Wahid, and J.~Davidson.
\newblock Visual representations for semantic target driven navigation, 2019.

\bibitem[Chang et~al.(2020)Chang, Gupta, and Gupta]{chang2020semantic}
M.~Chang, A.~Gupta, and S.~Gupta.
\newblock Semantic visual navigation by watching youtube videos.
\newblock In \emph{NeurIPS}, 2020.

\bibitem[Liang et~al.(2021)Liang, Chen, and Song]{liang2021sscnav}
Y.~Liang, B.~Chen, and S.~Song.
\newblock Sscnav: Confidence-aware semantic scene completion for visual semantic navigation, 2021.

\bibitem[Wu et~al.(2018)Wu, Wu, Tamar, Russell, Gkioxari, and Tian]{wu2018learning}
Y.~Wu, Y.~Wu, A.~Tamar, S.~Russell, G.~Gkioxari, and Y.~Tian.
\newblock Learning and planning with a semantic model, 2018.

\bibitem[Wortsman et~al.(2019)Wortsman, Ehsani, Rastegari, Farhadi, and Mottaghi]{wortsman2019learning}
M.~Wortsman, K.~Ehsani, M.~Rastegari, A.~Farhadi, and R.~Mottaghi.
\newblock Learning to learn how to learn: Self-adaptive visual navigation using meta-learning, 2019.

\bibitem[Rajvanshi et~al.(2023)Rajvanshi, Sikka, Lin, Lee, Chiu, and Velasquez]{rajvanshi2023saynav}
A.~Rajvanshi, K.~Sikka, X.~Lin, B.~Lee, H.-P. Chiu, and A.~Velasquez.
\newblock Saynav: Grounding large language models for dynamic planning to navigation in new environments, 2023.

\bibitem[Dorbala et~al.(2023)Dorbala, au2, and Manocha]{dorbala2023embodied}
V.~S. Dorbala, J.~F. M.~J. au2, and D.~Manocha.
\newblock Can an embodied agent find your "cat-shaped mug"? llm-guided exploration for zero-shot object navigation, 2023.

\bibitem[Shah et~al.(2023)Shah, Equi, Osinski, Xia, Ichter, and Levine]{shah2023navigation}
D.~Shah, M.~Equi, B.~Osinski, F.~Xia, B.~Ichter, and S.~Levine.
\newblock Navigation with large language models: Semantic guesswork as a heuristic for planning, 2023.

\bibitem[Majumdar et~al.(2023)Majumdar, Aggarwal, Devnani, Hoffman, and Batra]{majumdar2023zson}
A.~Majumdar, G.~Aggarwal, B.~Devnani, J.~Hoffman, and D.~Batra.
\newblock Zson: Zero-shot object-goal navigation using multimodal goal embeddings, 2023.

\bibitem[Liu et~al.(2023{\natexlab{a}})Liu, Li, Wu, and Lee]{liu2023llava}
H.~Liu, C.~Li, Q.~Wu, and Y.~J. Lee.
\newblock Visual instruction tuning.
\newblock In \emph{NeurIPS}, 2023{\natexlab{a}}.

\bibitem[Liu et~al.(2023{\natexlab{b}})Liu, Li, Li, and Lee]{liu2023improvedllava}
H.~Liu, C.~Li, Y.~Li, and Y.~J. Lee.
\newblock Improved baselines with visual instruction tuning, 2023{\natexlab{b}}.

\bibitem[Yenamandra et~al.(2023)Yenamandra, Ramachandran, Yadav, Wang, Khanna, Gervet, Yang, Jain, Clegg, Turner, Kira, Savva, Chang, Chaplot, Batra, Mottaghi, Bisk, and Paxton]{homerobotovmm}
S.~Yenamandra, A.~Ramachandran, K.~Yadav, A.~Wang, M.~Khanna, T.~Gervet, T.-Y. Yang, V.~Jain, A.~W. Clegg, J.~Turner, Z.~Kira, M.~Savva, A.~Chang, D.~S. Chaplot, D.~Batra, R.~Mottaghi, Y.~Bisk, and C.~Paxton.
\newblock Homerobot: Open vocab mobile manipulation, 2023.

\bibitem[Puig et~al.(2023)Puig, Undersander, Szot, Cote, Partsey, Yang, Desai, Clegg, Hlavac, Min, Gervet, Vondrus, Berges, Turner, Maksymets, Kira, Kalakrishnan, Malik, Chaplot, Jain, Batra, Rai, and Mottaghi]{puig2023habitat3}
X.~Puig, E.~Undersander, A.~Szot, M.~D. Cote, R.~Partsey, J.~Yang, R.~Desai, A.~W. Clegg, M.~Hlavac, T.~Min, T.~Gervet, V.~Vondrus, V.-P. Berges, J.~Turner, O.~Maksymets, Z.~Kira, M.~Kalakrishnan, J.~Malik, D.~S. Chaplot, U.~Jain, D.~Batra, A.~Rai, and R.~Mottaghi.
\newblock Habitat 3.0: A co-habitat for humans, avatars and robots, 2023.

\bibitem[Szot et~al.(2021)Szot, Clegg, Undersander, Wijmans, Zhao, Turner, Maestre, Mukadam, Chaplot, Maksymets, Gokaslan, Vondrus, Dharur, Meier, Galuba, Chang, Kira, Koltun, Malik, Savva, and Batra]{szot2021habitat}
A.~Szot, A.~Clegg, E.~Undersander, E.~Wijmans, Y.~Zhao, J.~Turner, N.~Maestre, M.~Mukadam, D.~Chaplot, O.~Maksymets, A.~Gokaslan, V.~Vondrus, S.~Dharur, F.~Meier, W.~Galuba, A.~Chang, Z.~Kira, V.~Koltun, J.~Malik, M.~Savva, and D.~Batra.
\newblock Habitat 2.0: Training home assistants to rearrange their habitat.
\newblock In \emph{Advances in Neural Information Processing Systems (NeurIPS)}, 2021.

\bibitem[Savva et~al.(2019)Savva, Kadian, Maksymets, Zhao, Wijmans, Jain, Straub, Liu, Koltun, Malik, Parikh, and Batra]{habitat19iccv}
M.~Savva, A.~Kadian, O.~Maksymets, Y.~Zhao, E.~Wijmans, B.~Jain, J.~Straub, J.~Liu, V.~Koltun, J.~Malik, D.~Parikh, and D.~Batra.
\newblock Habitat: {A} {P}latform for {E}mbodied {AI} {R}esearch.
\newblock In \emph{Proceedings of the IEEE/CVF International Conference on Computer Vision (ICCV)}, 2019.

\end{thebibliography}

\newpage

\section{SUPPLEMENTARY MATERIAL}
\subsection{SHORT TERM MEMORY}
When the agent chooses a node in the 3D scene representation to explore, It starts storing frame-wise data (RGBD image and processed segments from the open vocabulary object detector). If the agent detects the presence of the target object in a frame, then it finds all other views stored in the memory, from which this segment of the target object was visible. This is done by computing the spatial similarity (overlap between the pointcloud of detected object and all the segments in the memory). A LVLM is then tasked to reason the presence of the target object in all the retrieved frames. We use this reasoning to decide whether it is true or false detection. When the agent chooses a new node to explore, the current data stored in the short term memory is cleared.
\begin{figure}[!htbp]
    \centering
    \includegraphics[width=8cm]{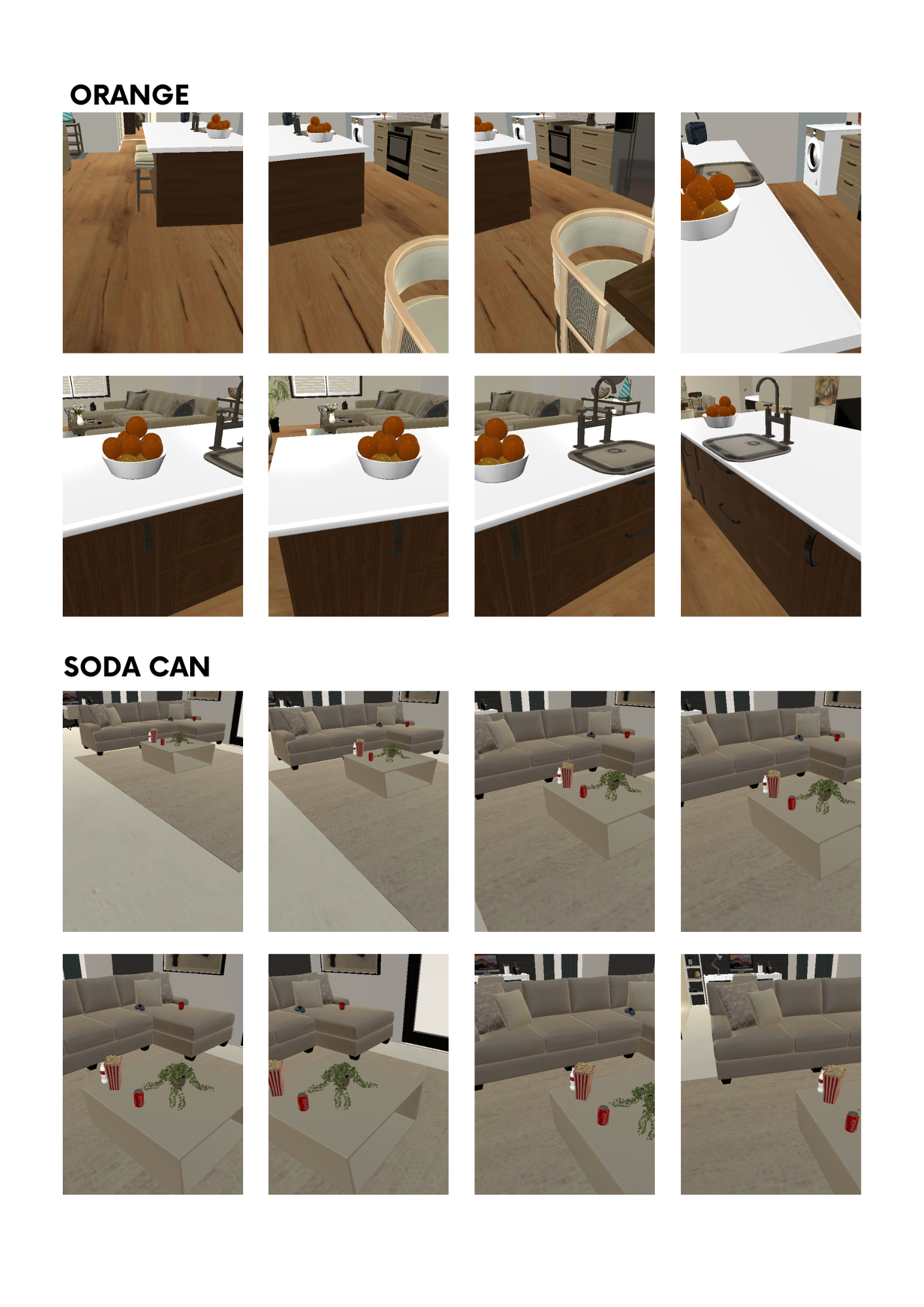}
    \caption{Examples of frames retrieved from the short term memory module, in which the target object is detected. The top 8 frames corresponds to those in which the target object orange was visible and the bottom 8 frames have the target object soda can visible in them.}
    \label{fig:stm}
\end{figure}
\subsection{PRUNING}
The pruner decides the abstraction level in which the planning occurs. For example, suppose we need to find an apple in an unseen environment. We can choose to represent the environment around us in different levels of abstraction, from room name to small objects like book. This representation will affect the quality of plans that we make to find apple. The task of the pruner is to provide an object level abstraction for the LLM based agent to plan a task. The pruner retains objects like computer table, chair, kitchen table, bed, cabinet and so on. To find an object like apple, the agent can choose to explore closer to one of these objects, in this case the kitchen table. Pruning is done with the help of incontext learning in LLMs. We provide a series of examples to LLM, that has a list of input objects and pruned objects and tasks LLM to do the same with a new list of input objects.

\begin{figure}[!htbp]
    \centering
    \includegraphics[width=15cm]{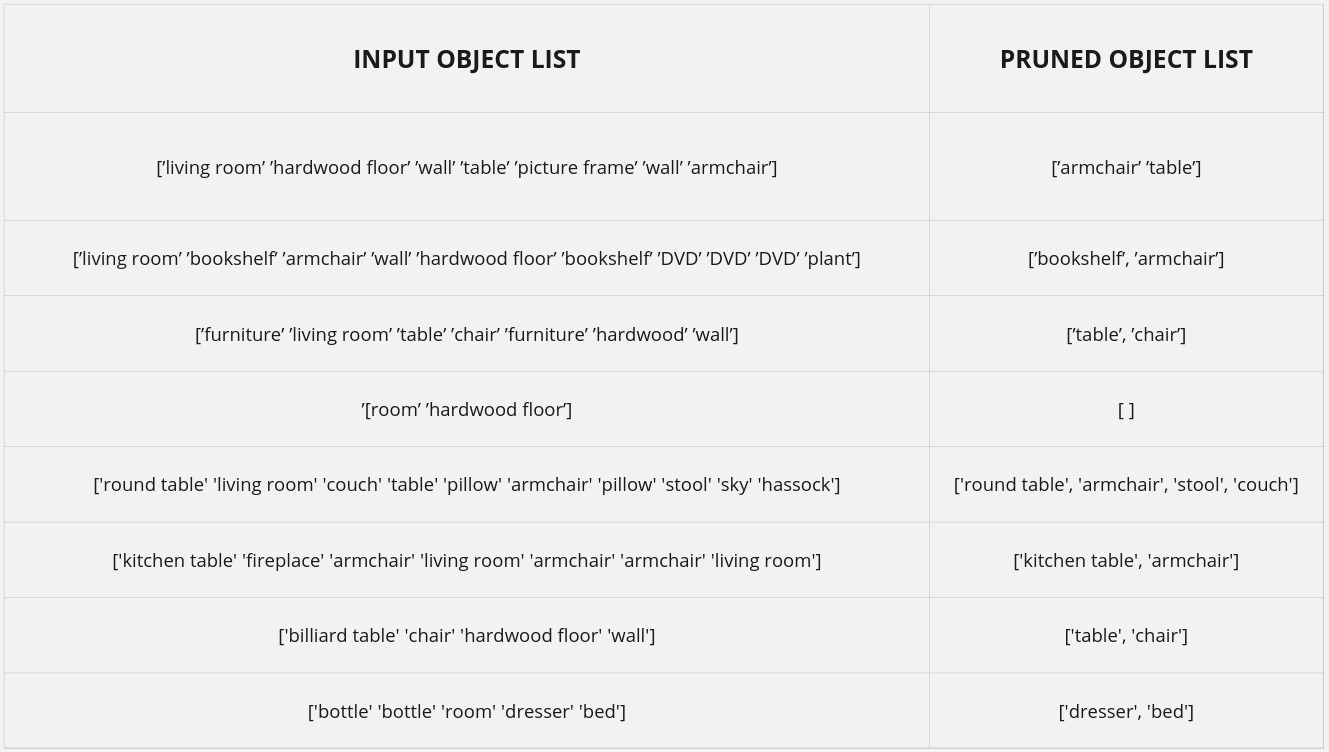}
    \caption{Input object list to LLM and the corresponding pruned list. The pruner here uses GPT-3.5 Turbo}
    \label{fig:prn_llm}
\end{figure}
        
\subsection{LLM BASED PLANNER}
The table below represents the plan executed by the agent (GPT-4 based agent) in an unseen environment to find an orange. Column 1 represents the object list along with the captions generated by LLaVA. Column two displays the frame when the agent identified a potential object (marked with a bounding box) to explore, for finding the target object. Column 3 has the frame at the end of exploring and Column 4 shows the agent's decision regarding the target object's presence.
\begin{figure}[!htbp]
    \centering
    \includegraphics[width=15cm]{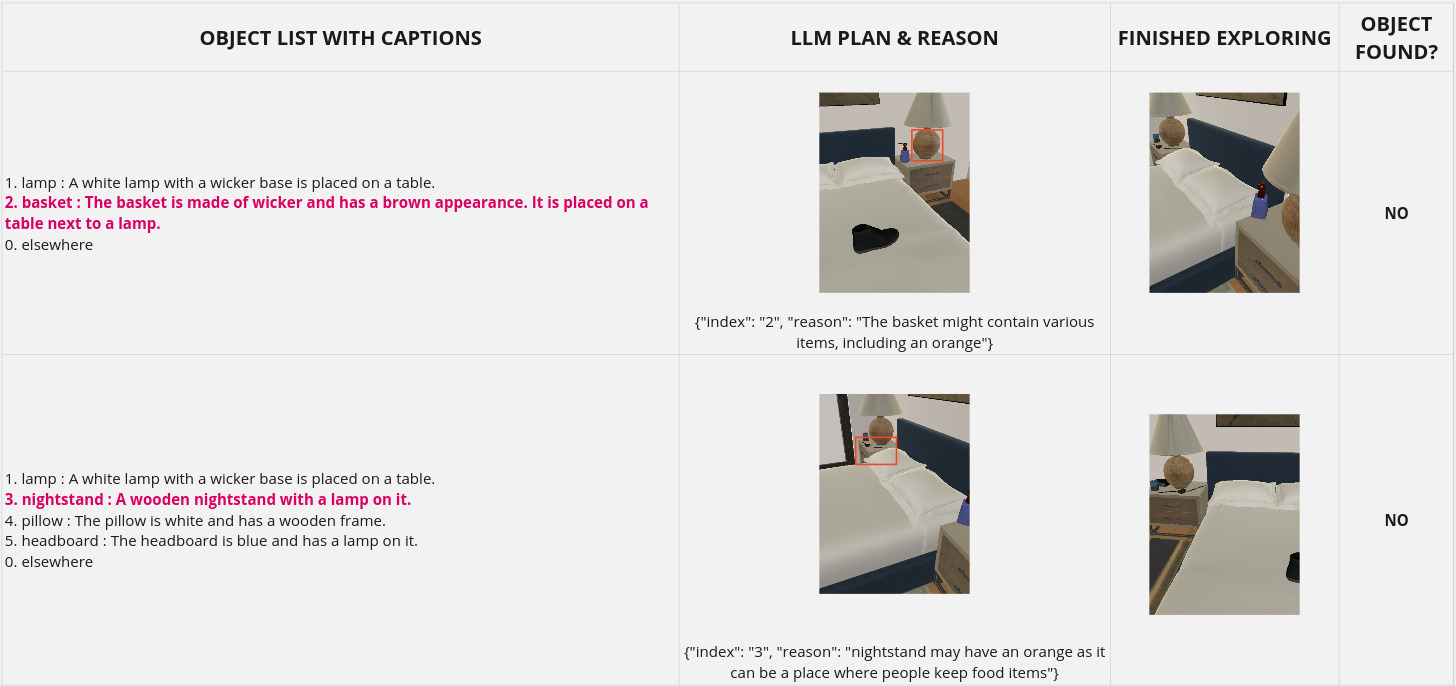}
    \label{fig:prn_llm}
\end{figure}

\begin{figure}[t]
    \includegraphics[width=15cm]{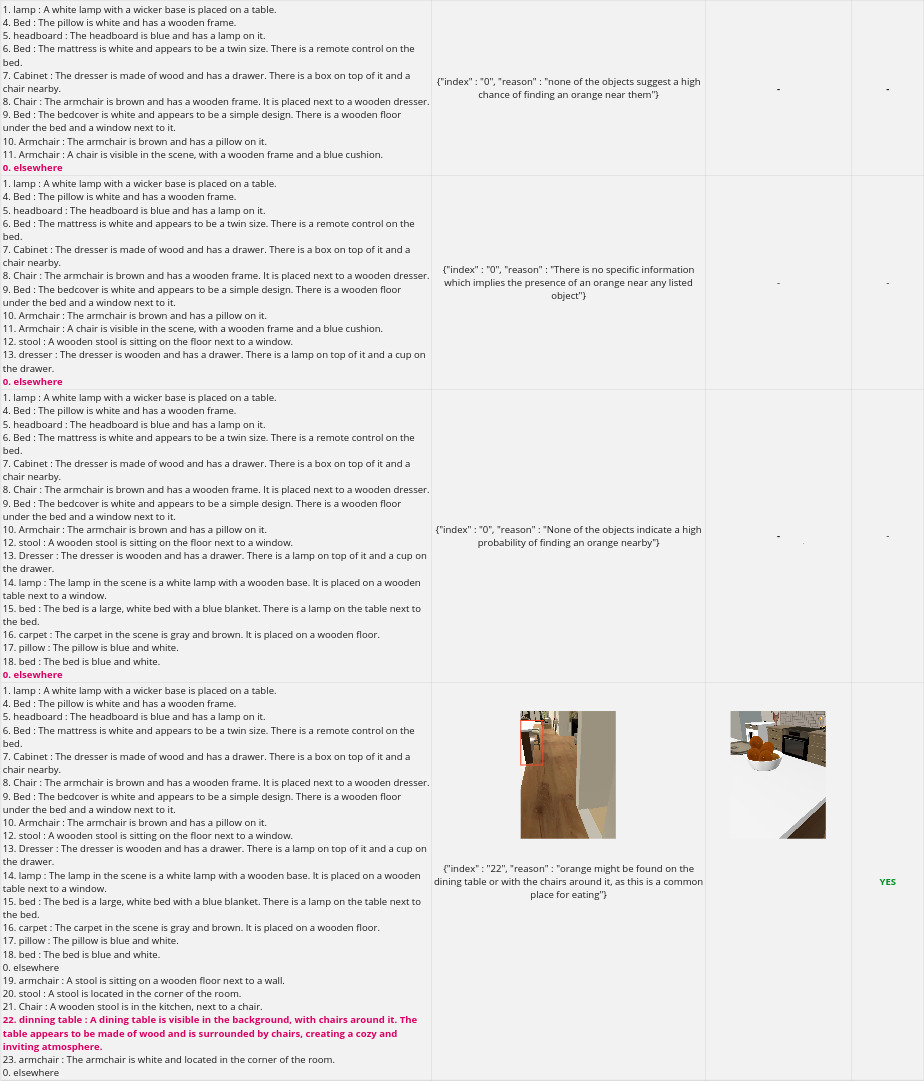}
    \label{fig:prn_llm}
\end{figure}

\end{document}